%% file: main.tex
\documentclass{article}
\usepackage{ijcai21}

\usepackage{times}
\usepackage{soul}
\usepackage{url}
\usepackage{hyperref}
\usepackage[utf8]{inputenc}
\usepackage[small]{caption}
\usepackage{graphicx}
\usepackage{amsmath}
\usepackage{amsthm}
\usepackage{booktabs}
\usepackage{algorithm}
\usepackage{subcaption}

\usepackage{vector}
\usepackage{amssymb,amsfonts}
\usepackage{xcolor}
\usepackage[capitalize]{cleveref}
\usepackage{siunitx}
\usepackage[noend]{algpseudocode}

\usepackage[c2]{optidef}

\newcommand{\ra}[1]{\renewcommand{\arraystretch}{#1}} 

\newcolumntype{R}{>{$}r<{$}}
\usepackage{todonotes}

\usepackage[style=authoryear,maxbibnames=10,maxcitenames=2,mincitenames=1,natbib=true,giveninits,dashed=false]{biblatex}
\usepackage{pgfplots}
\pgfplotsset{compat=newest}
\usepgfplotslibrary{groupplots}
\usepgfplotslibrary{dateplot}

\AtBeginDocument{ 
  \addtolength\abovedisplayskip{-0.25\baselineskip}%
  \addtolength\belowdisplayskip{-0.25\baselineskip}%
 \addtolength\abovedisplayshortskip{-0.25\baselineskip}%
 \addtolength\belowdisplayshortskip{-0.25\baselineskip}%
}
\usepackage[switch]{lineno} 

\title{Multi-Vehicle Control in Roundabouts using \\Decentralized Game-Theoretic Planning}

 \author{
 {Arec Jamgochian\and
 Kunal Menda\And
 Mykel J. Kochenderfer}
 \affiliations
 Department of Aeronautics and Astronautics, Stanford University
 \emails
 \{arec, kmenda, mykel\}@stanford.edu
 }
 
\begin{document}

\maketitle

\begin{abstract}
Safe navigation in dense, urban driving environments remains an open problem and an active area of research. Unlike typical predict-then-plan approaches, game-theoretic planning considers how one vehicle's plan will affect the actions of another. Recent work has demonstrated significant improvements in the time required to find local Nash equilibria in general-sum games with nonlinear objectives and constraints. 
When applied trivially to driving, these works assume all vehicles in a scene play a game together, which can result in intractable computation times for dense traffic. 
We formulate a decentralized approach to game-theoretic planning by assuming that agents only play games within their observational vicinity, which we believe to be a more reasonable assumption for human driving. Games are played in parallel for all strongly connected components of an interaction graph, significantly reducing the number of players and constraints in each game, and therefore the time required for planning. 
We demonstrate that our approach can achieve collision-free, efficient driving in urban environments by comparing performance against an adaptation of the Intelligent Driver Model and centralized game-theoretic planning when navigating roundabouts in the INTERACTION dataset. Our implementation is available at \href{http://github.com/sisl/DecNashPlanning}{http://github.com/sisl/DecNashPlanning}.
\end{abstract}

\input{1-intro}
\input{2-background}
\input{3-method}

\input{4-experiments}

\input{5-conclusion}

\section*{Acknowledgments}
This material is based upon work supported by the National Science Foundation Graduate Research Fellowship Program under Grant No. DGE-1656518. Any opinions, findings, and conclusions or recommendations expressed in this material are those of the author(s) and do not necessarily reflect the views of the National Science Foundation. 
This work is also supported by the COMET K2---Competence Centers for Excellent Technologies Programme of the Federal Ministry for Transport, Innovation and Technology (bmvit), the Federal Ministry for Digital, Business and Enterprise (bmdw), the Austrian Research Promotion Agency (FFG), the Province of Styria, and the Styrian Business Promotion Agency (SFG). 
The authors thank Simon Le Cleac’h, Brian Jackson, Lasse Peters, and Mac Schwager for their insightful feedback.



\printbibliography

\end{document}

%% file: 1-intro.tex
\section{Introduction} \label{sec:intro}

A better understanding of human driver behavior could yield better simulators to stress test autonomous driving policies.
A well-studied line of work attempts to build interpretable longitudinal or lateral motion models parameterized by only a few parameters. For example, the Intelligent Driver Model (IDM) control law has five parameters that govern longitudinal acceleration~\citep{treiber2000congested,kesting2010enhanced}. There are different ways to infer model parameters from data~\citep{liebner2012driver,buyer2019interaction,bhattacharyya2020online}. However, one can typically only use these models in well-defined, simple scenarios (e.g. highway driving). Building models from data in complicated scenarios (e.g. roundabouts) remains an open area of research.

A general approach is to assume every driver acts to optimize their own objective, as is the case in a Markov Game. If no agent has any incentive for unilaterally switching its policy, then the resulting joint policy is a Nash equilibrium~\citep{kochenderfer2021algorithms}. While typical policies may be formed using predictions about other vehicles, Nash equilibrium policies have the benefit of considering how agents may respond to the actions of others. Nash equilibrium (and other game-theoretic) policies therefore can more closely capture human driving, where drivers reason about how their driving decisions will affect the plans of other drivers and negotiate accordingly.

Nash equilibrium strategies for finite-horizon linear-quadratic games obey coupled Riccati differential equations, and can be solved for in discrete time using dynamic programming~\citep{bacsar1998dynamic,engwerda2007algorithms}. More recent work has explored efficiently finding Nash equilibria in games with non-linear dynamics by iteratively linearizing them~\citep{fridovich2020efficient} or by searching for roots of the gradients of augmented Lagrangians~\citep{cleac2019algames}. Both kinds of approaches have been applied to urban driving scenarios. However, they focus on scenarios where the number of agents is low, and the methods are empirically shown in the latter work to scale poorly as the number of agents is increased.

\begin{figure}
    \centering
    \includegraphics[width=0.8\columnwidth]{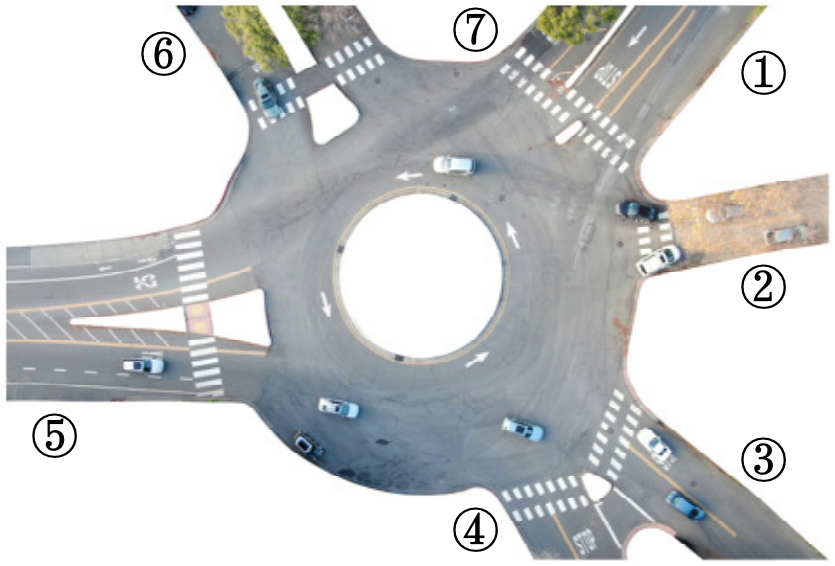}
    \caption{The \textit{DR\_USA\_Roundabout\_FT} roundabout~\citep{zhan2019interaction}. Planning in a dense traffic environment requires negotiating with multiple agents, but not with all agents in a scene.}
    \label{fig:drchn}
\end{figure}

In dense, urban driving environments, the number of agents in a particular scene can be quite large. Navigating a roundabout, such as the one shown in~\Cref{fig:drchn}, can require successfully negotiating with many other drivers. Luckily, modeling multi-agent driving in dense traffic as one large, coordinated game of negotiations is unrealistic. Drivers typically perform negotiations with the only vehicles closest to them. Furthermore, they cannot negotiate with drivers who do not see them. 

In this work, we formulate multi-vehicle decentralized, game-theoretic control under the latter assumption that games are only played when drivers see one another. We use a time-dependent interaction graph to encodes agents' observations of other agents, and we delegate game-theoretic control to strongly connected components (SCCs) within that graph. Furthermore, we make the assumption any time a driver observes a vehicle that does not observe them in turn, that driver must use forecasts of the other vehicle's movements in their plan. 
Thus, the game of an SCC of agents has additional collision constraints imposed by outgoing edges from that SCC.

We demonstrate our method on a roundabout simulator derived from the INTERACTION dataset~\citep{zhan2019interaction}. We observe that a decentralized model based on IDM is untenable for safe control, and that our decentralized approach to game-theoretic planning can generate safe, efficient behavior. We observe that our approach provides significant speedup to centralized game-theoretic planning, though it is still not fast enough for real-time control. 

In summary, our contributions are to:
\begin{itemize}
    \item Introduce a method for achieving (partially) decentralized game-theoretic control, and
    \item Apply that method to demonstrate human-like driving in roundabouts using real data.
\end{itemize}
The remainder of this paper is organized as follows: \cref{sec:back} outlines background material, \cref{sec:method} describes our approach and methodologies, \cref{sec:experiments} highlights experiments and results, and \cref{sec:conclusion} concludes.

%% file: 2-background.tex
\section{Background} \label{sec:back}

Research to model and predict human driver behavior was recently surveyed by~\citet{brown2020modeling}. 
The field of imitation learning (IL) seeks to learn policies that mimic expert demonstrations. One class of IL algorithms requires actively querying an expert to obtain feedback on the policy while learning. Naturally, this method does not scale~\citep{ross2010efficient,ross2011reduction}. 

Other methods use fixed expert data and alternate between optimizing a reward function, and re-solving a Markov decision process for the optimal policy, which can be expensive~\citep{abbeel2004apprenticeship,ziebart2008maximum}.
More recent methods attempt to do these steps progressively while using neural network policies and policy gradients~\citep{finn2016guided,ho2016model}. Generative Adversarial Imitation Learning (GAIL) treats IL as a two-player game in which a generator is trying to generate human-like behavior to fool a discriminator, which is, in turn, trying to differentiate between artificial and expert data~\citep{ho2016generative}. GAIL was expanded to perform multi-agent imitation learning in simulated environments~\citep{song2018multi}. GAIL has recently been tested with mixed levels of success on real-world highway driving scenarios~\citep{bhattacharyya2018multi,bhattacharyya2020modeling}. GAIL has not yet been implemented in more complex driving environments (e.g. roundabouts), where success will likely require significant innovation.

The literature on multi-agent planning is rich and extensive. 
A common assumption is cooperation, which can be modeled with one single reward function that is shared by all agents. 

Multi-agent cooperative planning is surveyed recently by~\cite{torreno2017cooperative}.
In real driving, agents do not cooperate, but rather optimize their individual reward functions. One approach is to model real driving as a general-sum dynamic game. In non-cooperative games, different solution concepts have been proposed. Nash equilibria are solutions in which no player can reduce their cost by changing their strategy alone~\citep{kochenderfer2021algorithms}. The problem of finding Nash equilibria for multi-player dynamic games in which player strategies are coupled through joint state constraints (e.g. for avoiding collisions) is referred to as a Generalized Nash Equilibrium Problem (GNEP)~\citep{facchinei2007generalized}.

Nash equilibrium strategies for finite-horizon linear-quadratic games obey coupled Riccati differential equations, and can be solved for in discrete time using dynamic programming~\citep{bacsar1998dynamic,engwerda2007algorithms}. A recent approach, iLQGames, can solve more general differential dynamic games by iteratively making and solving linear-quadratic approximations until convergence to a local, closed-loop Nash equilibrium strategy~\citep{fridovich2020efficient}. An alternative method, ALGAMES, uses a quasi-Newton root-finding algorithm to solve a general-sum dynamic game for a local open-loop Nash equilibrium~\citep{cleac2019algames}. Unlike iLQGames, ALGAMES allows for hard constraints in the joint state and action space, but it must be wrapped in model-predictive control to be used practically. 
We note that neither method finds global Nash equilibria, and the same game can result in different equilibria depending on the initialization. The problem of inferring which Nash equilibrium most closely matches observations of other vehicles is considered by~\cite{peters2020inference}. 
Since the reward functions of other vehicles may be unknown, additional work formulates an extension in which an Unscented Kalman Filter is used to update beliefs over parameters in the other vehicle's reward functions~\citep{cleac2020lucidgames}.

In this paper, we adopt the convention presented in ALGAMES. We denote the state and control of player $i$ at time $t$ as $x^i_t$ and $u^i_t$ respectively. We denote the full trajectory of all agents $x_{1:T}^{1:N}$ as $X$ for shorthand. $X^i$ indicates agent $i$'s full trajectory, $X_t$ indicates the joint state at time $t$. Similarly, we use $U$ to indicate the joint action plan $u_{1:T}^{1:N}$, $U^i$ to indicate agent $i$'s full plan, and $U_t$ to indicates the joint action at time $t$. The cost function of each player is denoted $J^i(X,U^i)$. 

The objective of each player is to choose actions that minimize the cost of their objective function subject to constraints imposed by the joint dynamics $f$, and by external constraints $C$ (e.g. collision avoidance constraints, action bounds, etc.):
\begin{mini}|s|
{\substack{X, U^i}}
{J^i(X,U^i)}
{\label{eq:game}}
{}
\addConstraint{X_{t+1}}{= f(t, X_t, U_t)}{, t = 1,\ldots,T-1}
\addConstraint{C(X,U)}{\leq 0}
\end{mini}
A joint plan $U$ optimizes the GNEP and is, therefore, an open-loop Nash equilibrium if each plan $U^i$ is optimal to its respective problem when the remaining agents plans $U^{-i}$ are fixed. That is to say, no agent can improve their objective by only changing their own plan. 

%% file: 3-method.tex
\section{Methodology} \label{sec:method}

\subsection{Decentralized Game-Theoretic Planning} \label{sec:decplan}

We note that in dense traffic, the assumption that all agents play a game with all other agents is unrealistic. More realistically, agents only negotiate with a subset of the agents in the scene at any one given point in time. If there are many agents in the scene, then agents typically negotiate with the ones in their immediate vicinity.

Furthermore, this game-theoretic negotiation exists only when there is two-way communication between agents. If two drivers can observe one another, they can negotiate, but if a trailing vehicle is in a leading driver's blind spot, then the leading driver is agnostic to the trailing vehicle's actions while the trailing driver must plan around predictions about the behavior of the leading vehicle.

We encode this information flow using a directed graph $\mathcal{G} = (\mathcal{V}, \mathcal{E})$ where node $i \in \mathcal{V}$ captures information about agent $i$ and an edge $(i,j) \in \mathcal{E}$ captures whether agent $i$ can observe agent $j$. We define the agent $i$'s neighbor set $\mathcal{N}_i = \{j \mid (i,j) \in \mathcal{E}\}$ as the set of all agents that agent $i$ actively observes.

At each point in time, we can generate the interaction graph $\mathcal{G}_t$ based on the agents each agent actively observes. We assume that a human driver can only reason about a small subset of agents that are within its field of view, so we add those observed agents to the ego vehicle's neighbor set. The cardinality of each neighbor set $|\mathcal{N}_i|$ can therefore be much smaller than the total number of agents in a scene. We then instantiate one game for each of the strongly connected components (SCCs) of $\mathcal{G}_t$.

\begin{figure}
    \centering
    \includegraphics[width=3in]{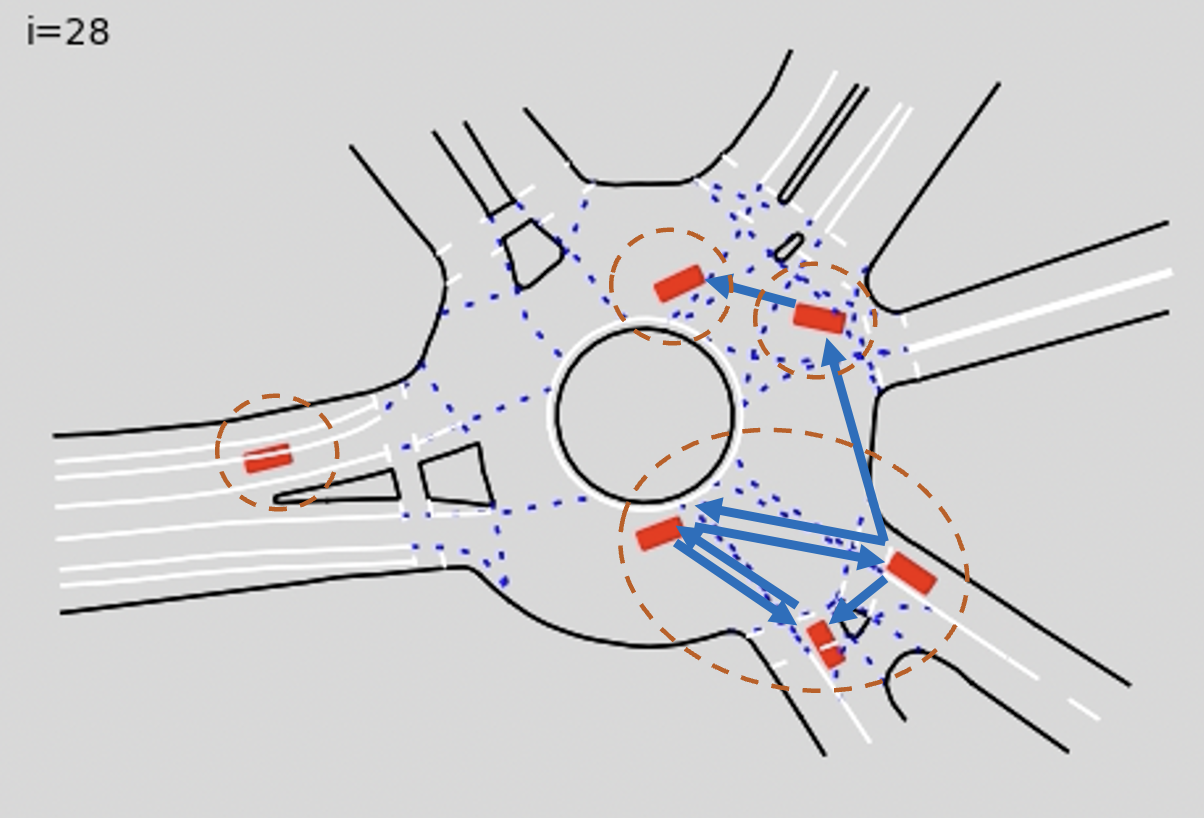}
    \caption{Sample roundabout frame from the INTERACTION dataset~\citep{zhan2019interaction}. Observation edges are denoted with arrows, and strongly connected components are drawn with circles. Each strongly connected component must make a game-theoretic plan while respecting forecasts of agents from edges leaving that component.}
    \label{fig:roundabout}
\end{figure}

Within each SCC, agents must compute a Nash equilibrium. Agents within the SCC negotiate with one another, but are constrained by the predictions of observed agents who do not observe them (e.g. the outgoing edges of each strongly connected component). If two agents observe one another, and one of those two agents observed a third leading car, then the agents must make a joint plan subject to predictions about the third car. This can be done by instantiating a game with all three cars and fixing the actions of the third car to be those of its forecasted plan. In our experiments, we forecast that observed agents maintain their velocities, however, our method is amenable to more complex forecasting methods. An example interaction graph is shown in ~\Cref{fig:roundabout}. 

To control the vehicles, this decentralized planning is wrapped in model predictive control. Once each planner determines the plan of all the agents in each sub-component, the first action of each plan is taken, a new interaction graph is generated, and new game-theoretic planners are instantiated. Decentralized game-theoretic planning is summarized in \Cref{alg:decnash}.

\begin{algorithm}
\caption{Decentralized Game-Theoretic Planning}\label{alg:decnash}
\begin{algorithmic}[1]
\Procedure{DecNash}{$X_1$}
\For{$t \gets 1:T$}
\State $g_t\gets \text{InteractionGraph}(X_t)$
\State $SCCs \gets \text{StronglyConnectedComponents}(g_t)$
\For{$SCC \in SCCs$}
\State $game\gets \text{Game}(SCC, X_t)$
\State $U[SCC.nodes] \gets \text{NashSolver}(game)$
\EndFor\label{euclidendwhile}
\State $X_{t+1} \gets \text{ExecuteAction}(X_{t}, U)$
\EndFor
\EndProcedure
\end{algorithmic}
\end{algorithm}

We believe that this model of human driving more realistically captures real-life driving negotiations, where each driver makes plans based only on a few observed neighboring vehicles, and where game-theoretic negotiations only occur between vehicles that actively observe one other. For example, if a driver observes a vehicle they know is not observing them, then that driver must plan using forecasts of the observed vehicle's motion, rather than play a game to communicate their intent. Furthermore, this decentralized approach is much more scalable since the number of players, as well as the $\mathcal{O}(n^2)$ safety constraints within each game, stays relatively small even when there are many vehicles in the driving scenario. The dependence on the number of players was observed empirically by~\citet{cleac2019algames}, who saw a ten- to twenty-fold increase in computation time for their method when moving from two to four agents per game.

\subsection{Roundabout Game} \label{sec:roundabout}

We formulate the local negotiations between agents in a roundabout as separate GNEPs to be solved in parallel. Each problem can be thought of to have $N$-players, $N_c$ of which are controlled, and $N_o=N-N_c$ of which are observed. $N_c$ is the size of the SCC being considered, and $N_o$ is the number of outgoing edges from that SCC. Since we only control $N_c$ agents in our decentralized plan, we only need to consider the first $N_c \leq N$ optimization problems in the dynamic game in~\cref{eq:game}. 

In our roundabout game, we assume that vehicle trajectories are determined beforehand (e.g. by a trajectory planner), and the goal of our planner is to prescribe accelerations of vehicles along their respective trajectories. 
We do so by parametrizing a vehicle's trajectory using a polynomial mapping a distance travelled along the path to the $x$ and $y$ positions of the vehicle.
A vehicle modulates acceleration along this pre-determined path in order to safely and efficiently navigate to its goal.
We think it is valid to assume that trajectories can be computed \textit{a priori} in roundabouts, but this may be a less-valid assumption in highway traffic, where agents can plan their paths to react to other agents. 
Our decentralized method can, however, be extended to other driving environments by reformulating the game setup.

With our assumptions, each vehicle's state can be written:
\begin{equation}
    x^i_t = \begin{bmatrix}
    p_{xt}^i = p(s^i_t;\theta_{x^i}) \\ p_{yt}^i = p(s^i_t;\theta_{y^i}) \\ s^i_t \\ v^i_t
\end{bmatrix}\text{, where }
p(s;\theta) = \sum_{i=0}^{|\theta|-1} \theta_i s^i\text{.}
\end{equation}
Here, $s$ and $v$ encode the vehicle position and velocity along its path and are the minimal state representation. To simplify collision constraints, the first two elements in the state expand the path position to $x$ and $y$ locations using coefficients $\theta_{x^i}$ and $\theta_{y^i}$. Each vehicle directly controls its own acceleration, leading to the following continuous-time dynamics, which are discretized at solve time:

\begin{equation}
    \dot{x}^i_t = \begin{bmatrix}
    \dot{p}(s^i_t, v^i_t;\theta_{x^i}) \\ \dot{p}(s^i_t, v^i_t;\theta_{y^i}) \\ v^i_t \\ u^i_t
\end{bmatrix}\text{, where }
\dot{p}(s,v;\theta) = v\sum_{i=1}^{|\theta|-1} \theta_i is^{i-1}\text{.}
\end{equation}

To incentivize vehicles to reach their velocity targets while penalizing unnecessary accelerations, we use a linear-quadratic objective:

\begin{equation}
    J^i(X^i, U^i) = \sum_{t=1}^T q^i\|v^i_t-v^i_\text{target}\|_2^2 + r^i\|u^i_t\|_2^2\text{,}
\end{equation}
where $q_i$ and $r_i$ are vehicle-dependent objective weighting parameters, and $v^i_\text{target}$ is the vehicle's desired free-flow speed. 

Controls are constrained to be within the control limits of each controlled vehicle, and to be equivalent to the predicted controls for the observed vehicles. That is, for all $t\in \{1,...,T\}$, we have 
\begin{align}
    u^i_\text{min} \leq u^i_t \leq u^i_\text{min} & \quad\forall i \in \{1,...,N_c\}\\
    u^i_t=\hat{u}^i_t &\quad\forall i \in \{N_c+1,...,N\} \label{eq:equality}
\end{align}
where $\hat{u}^i_t$ is the predicted control of observed vehicle $i$ at time $t$.

Finally, we impose collision constraints between each pair of (both observed and unobserved) vehicles at every point in time. That is, for every pair of agents $i$ and $j$ at every time in the horizon, we introduce constraints
\begin{equation}
    (p_{xt}^i-p_{xt}^j)^2+(p_{yt}^i-p_{yt}^j)^2 \geq r_\text{safe}^2\text{,}
\end{equation}
where $r_\text{safe}$ is a safety radius.

%% file: 4-experiments.tex
\section{Experiments} \label{sec:experiments}

In our experiments, we empirically demonstrate our approach's ability to generate human-like behavior in real-world, complicated driving scenarios. We use the INTERACTION dataset~\citep{zhan2019interaction}, which consists of drone-collected motion data from scenes with complicated driving interactions. We focus on control in a roundabout, a traditionally difficult driving scenario that requires negotiating with potentially many agents. From the data, we extract vehicle spawn times and approximate true paths with twentieth-order polynomials. In our simulator, we fix these values and allow for different methods to provide (one-dimensional) acceleration control for all agents in the scene, thereby controlling agents along paths generated from real-world data. Our code is made available at \texttt{github.com/sisl/DecNashPlanning}.

We analyze behavior generated from a fully decentralized policy in which vehicles follow an adaptation of IDM~\citep{treiber2000congested}, a centralized game-theoretic policy, and our decentralized game-theoretic method. We discuss observed behavior and compare collision rates, average speed shortfalls, planning times, and agents in each game.

\subsection{IDM Adaptation}

Given speed $v(t)$ and relative speed $r(t)$ to a followed vehicle, the Intelligent Driver Model (IDM) for longitudinal acceleration predicts acceleration as 
\begin{align}
    d_{\text{des}} &= d_\text{min} + \tau \cdot v(t) - \frac{v(t) \cdot r(t)}{2\sqrt{a_\text{max} \cdot b_\text{pref}}} \\ 
    a(t) &= a_\text{max} \left[1-\left(\frac{v(t)}{v_\text{target}}\right)^4-\left(\frac{d_\text{des}}{d(t)}\right)^2\right]\text{,}
\end{align}
where $d_\text{des}$ is the desired distance to the lead vehicle, $d_\text{min}$ is the minimum distance to the lead vehicle that the driver will tolerate, $\tau$ is the desired time headway, $a_\text{max}$ is the controlled vehicle’s maximum acceleration, $b_\text{pref}$ is the preferred deceleration, and $v_\text{target}$ is the driver’s desired free-flow speed.

Though IDM is intended for modeling longitudinal driving, we adapt it for use in roundabout driving by designating the closest vehicle within a $20^{\circ}$ half-angle cone from the controlled vehicle's heading as the ``follow'' vehicle, and calculating the relevant parameters by casting the ``follow'' vehicle to the controlled vehicle's path.

\subsection{Game-Theoretic Planning}

We implement our decentralized game-theoretic planning approach as described in \Cref{sec:decplan}. To generate an interaction graph, we say agent $i$ observes agent $j$ (e.g. $(i,j) \in \mathcal{E}$) if $j$ is within a $20$ meter, $120^\circ$ half-angle cone from $i$'s heading. To compute strongly connected components and their outgoing edges, we use Kosaraju's algorithm~\citep{nuutila1994finding}, which uses two depth-first search traversals to compute strongly connected components in linear time.

We implement the one-dimensional dynamic roundabout game setup as described in \Cref{sec:roundabout}. For predictions of the outgoing edges of each strongly connected component, we assume each observed uncontrolled vehicle will maintain its velocity throughout the planning horizon (i.e. $\hat{u}^i_t$ from~\cref{eq:equality} is 0 for all $t$). We use \textit{Algames.jl}~\citep{cleac2019algames} to solve for Nash equilibria by optimizing an augmented Lagrangian with the objective gradients. At each point in time, we plan a four-second horizon, where decisions are made every 0.2 seconds.

We also compare our method against a fully centralized approach---that is, one where all agents in a frame play a single game together.

\subsection{Results}

As a baseline, our IDM adaptation performs surprisingly well even though IDM is designed to model single-lane driving. We observe successful navigation within the first few hundred frames, with collisions happening rarely when two vehicles do not observe one another while merging. As time progresses, we observe severe slowdowns as more agents enter the scene, and an eventual standstill. During standstill, the number of collisions increases dramatically as new vehicles enter the scene before space is cleared. One could imagine designing sets of rules to circumvent problems from the IDM-based approach, but rule design for such a complex environment can be very difficult because of the edge cases that must be considered. Regardless, this approach highlights the lack of safety in simple systems.

On the contrary, decentralized game-theoretic planning generates human-like driving without explicitly encoding rules. We observe agents accelerating and decelerating naturally with the flow of locally observed traffic. We also observe instances of agents safely resolving complex traffic conflicts which would be difficult to encode in rules-based planners. Although our quantitative experiments focus on evaluating performance at a single, more difficult roundabout, we see human-like, safe driving emerge across roundabout environments. That is, we observe collision-free generalization to different roundabouts without doing any additional parameter tuning. For videos of resulting behavior please refer to supplementary material.

\begin{figure} 
    \centering
    \begin{subfigure}{0.45\textwidth}
        \centering
        \includegraphics[width=0.9\textwidth]{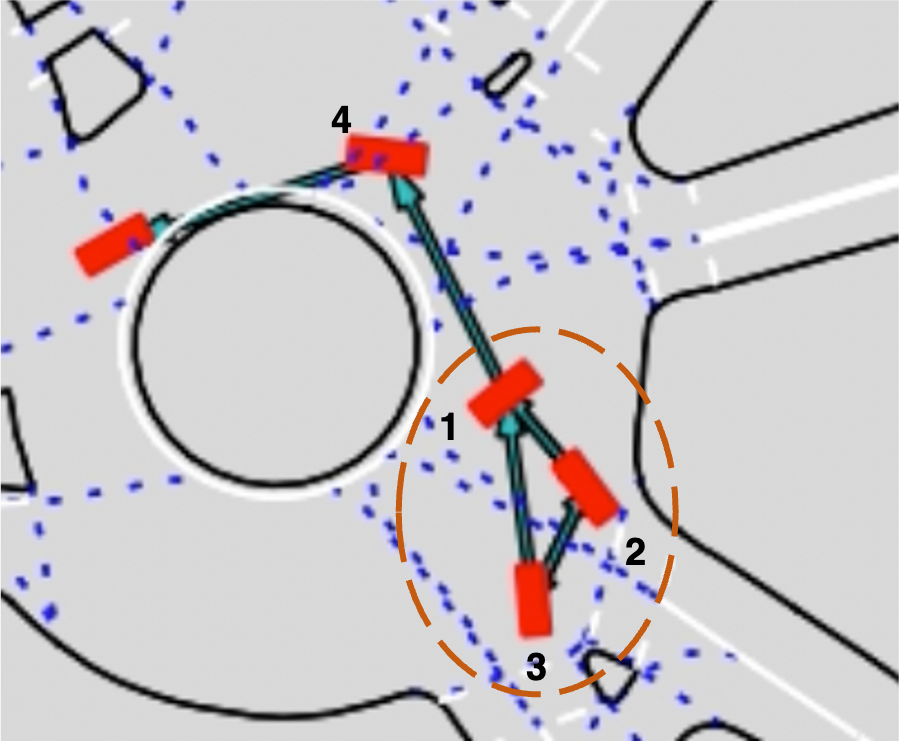} 
        \caption{Interaction Graph}
    \end{subfigure}\vfill
    \par\bigskip
    \begin{subfigure}{0.45\textwidth}
        \centering
        \centerline{\scalebox{0.95}{\input{figs/PlannedTrajs.tex}}}
        \caption{Planned Trajectory for SCC}
    \end{subfigure}\vfill
    \par\bigskip
    \begin{subfigure}{0.45\textwidth}
        \centering
        \centerline{\scalebox{0.9}{\input{figs/VelocityProf.tex}}}
        \caption{Planned Velocity Profile for SCC}
    \end{subfigure}
    \caption{A snapshot of an interaction graph, with a particular strongly connected component highlighted (a). There are three agents in the SCC and a fourth one which is observed but not controlled. The three vehicles inside the SCC safely negotiate with one another en route their goals. The planned Nash equilibrium trajectory is shown in (b) with points for every 0.2 seconds. The planned velocity profile is depicted in (c).}
    \label{fig:decnash}
\end{figure}

A sample decentralized open-loop, Nash equilibrium plan can be seen in \Cref{fig:decnash}. In it, one highlighted strongly connected component consists of three vehicles that observe one another, and observe one other vehicle that does not observe them in turn. The optimal plan for this strongly connected component has the three vehicles in the component safely interacting to proceed towards their goals, while the other observed but non-controlled vehicle maintains its velocity.

\begin{figure*} 
    \centering
    \begin{subfigure}{0.331\textwidth}
        \centering
        \includegraphics[width=\textwidth]{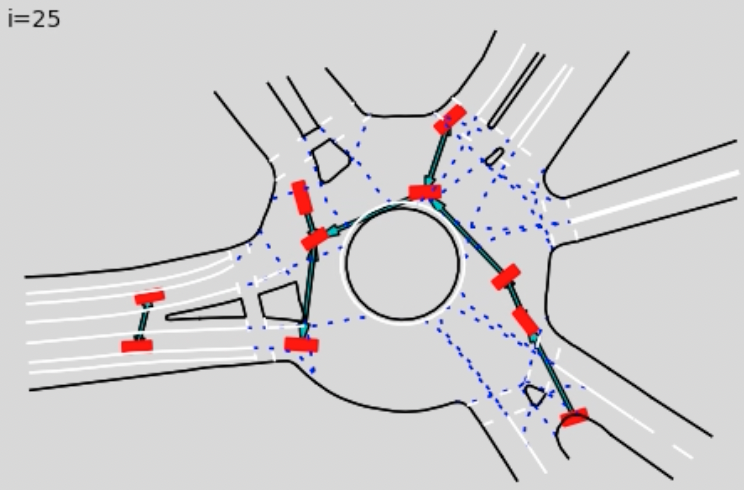} 
    \end{subfigure}\hfill
    \begin{subfigure}{0.331\textwidth}
        \centering
        \includegraphics[width=\textwidth]{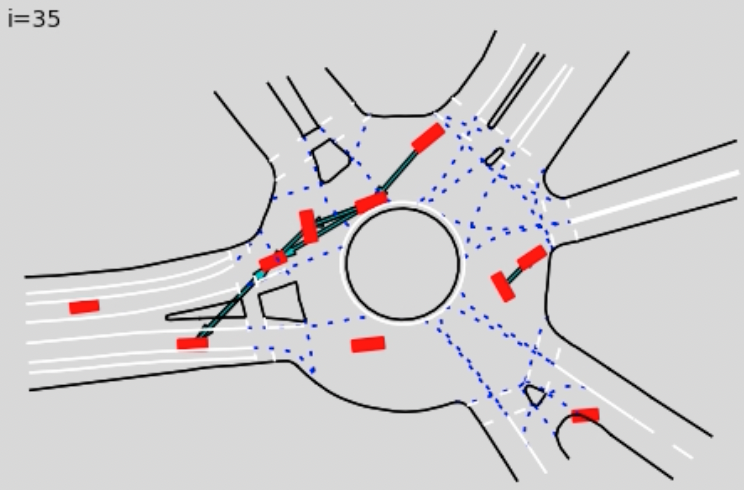} 
    \end{subfigure}\hfill
    \begin{subfigure}{0.331\textwidth}
        \centering
        \includegraphics[width=\textwidth]{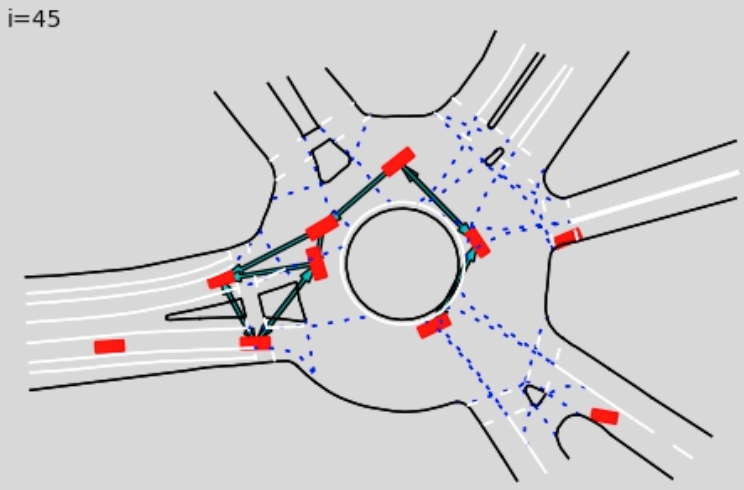} 
    \end{subfigure}
    \vfill
    \begin{subfigure}{0.331\textwidth}
        \centering
        \includegraphics[width=\textwidth]{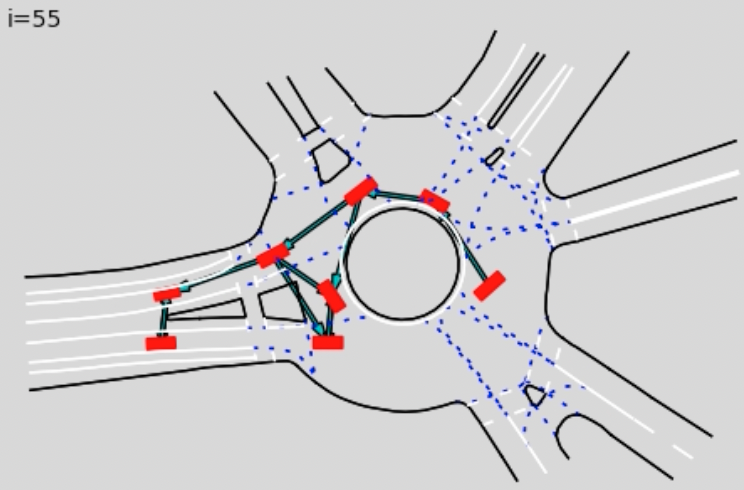} 
    \end{subfigure}\hfill
    \begin{subfigure}{0.331\textwidth}
        \centering
        \includegraphics[width=\textwidth]{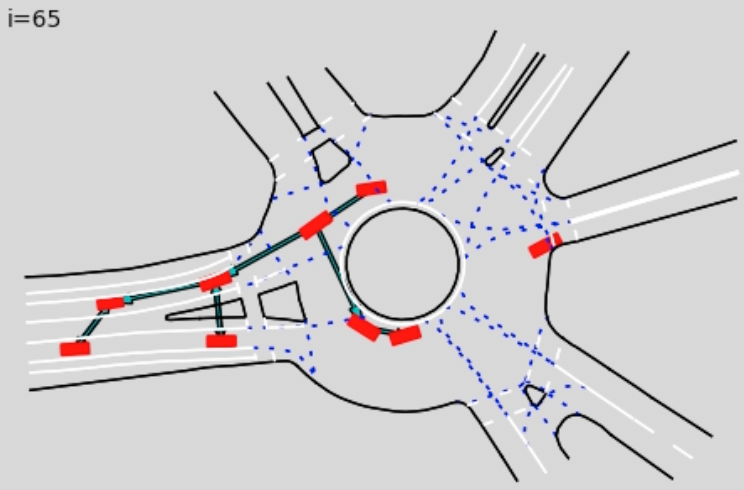} 
    \end{subfigure}\hfill
    \begin{subfigure}{0.331\textwidth}
        \centering
        \includegraphics[width=\textwidth]{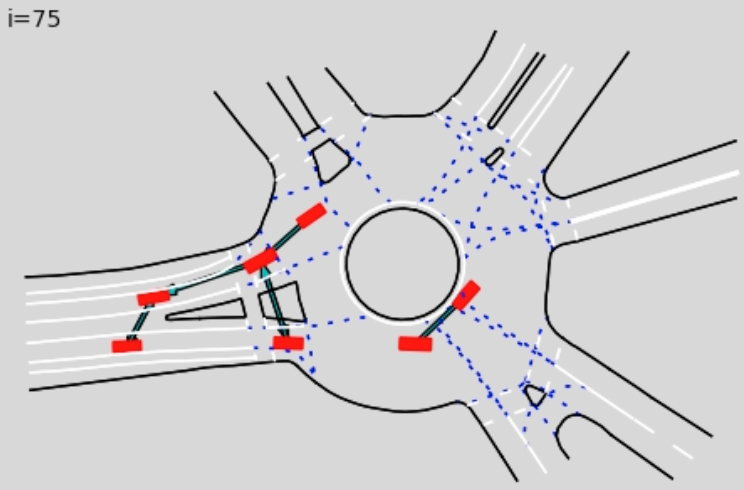} 
    \end{subfigure}
    \caption{A set of snapshots of traffic conflicts which gets resolved safely through decentralized game-theoretic planning. Snapshots are shown at one second intervals (ten decision frames). The vehicles quickly resolve right-of-way and merge safely without the need for a complex set of rules.}
    \label{fig:resolution}
\end{figure*}

\Cref{fig:resolution} depicts the safe resolution of a traffic conflict by our method that would be otherwise hard to resolve. In it, the vehicles resolve who has right-of-way and merge safely behind one another without wholly disrupting the flow of traffic, which happens often in the IDM adaptation baseline.

In~\Cref{table:results}, we compare the performance of the IDM adaptation, centralized game-theoretic planning, and decentralized game-theoretic planning in some key metrics averaged over five 100-second roundabout simulations where decisions are made at 10 Hz. Namely, we compare the number of collisions per 100 seconds, the average shortfall below target speed, the number of players per game given our choice of interaction graph, and the time to generate a joint policy. For our decentralized approach, we define the number of players per game and the time to generate a joint policy using the largest values across strongly connected components at each time step, as the games can be parallelized. We note that our solution times are slow because we are approximating the paths as twentieth-order polynomials, and the times are meant to highlight relative improvement when using a decentralized approach.

\begin{table}[htpb]

\ra{1.2}
\caption{A comparison of performance metrics with the IDM adaptation, centralized game-theoretic planning (C-Nash), and decentralized game-theoretic planning (Dec-Nash) models, performed over five scenes of 100 seconds of driving. We compare the mean number of collisions per 100 seconds (collisions), the mean vehicle shortfall below target speed ($v_\text{target}-v_\text{avg}$), the number of players in the largest game at each time step (\# players), and the time to generate a joint policy.}
\centering
\scalebox{0.82}{\input{datatable}}
\label{table:results}
\end{table}

We observe that though generating a joint policy is quick in the IDM adaptation, it leads to unsafe behavior that is significantly reduced in the game-theoretic approaches. We observe that the game-theoretic approaches are both a) much safer in terms of collision rate and b) much more efficient in terms of vehicle speed. We see an average of 55.8, 4.8, and 0.2 collisions per 100 seconds when using the IDM, centralized game-theoretic, and decentralized game-theoretic policies respectively. Similarly, we see that the mean vehicle shortfall velocity improves from 8.7 m/s with IDM, to 3.2 m/s with the centralized game-theoretic policy, to 2.8 m/s with our method.

We notice that using a centralized game-theoretic approach leads to more driving hesitancy, which ultimately results in slower average velocities when compared to our approach. We observe collisions in the centralized approach being caused by a) the solver being unable to find adequate solutions when the games are too large, and b) vehicles spawning before their path has been cleared. With the decentralized game-theoretic policy, we only observed one collision in the 500 total seconds of driving, which we believe to have also been caused by the game in question being too large.

Furthermore, we observe that our decentralized approach can generate plans much more quickly than a centralized one. We also expect that in larger urban intersections and roundabouts with much denser traffic, the performance gaps would become even more profound, and our method will be a necessity to keep game-theoretic plan generation tractable.

%% file: figs/PlannedTrajs.tex
\begin{tikzpicture}

\definecolor{color0}{rgb}{0.12156862745098,0.466666666666667,0.705882352941177}
\definecolor{color1}{rgb}{1,0.498039215686275,0.0549019607843137}
\definecolor{color2}{rgb}{0.172549019607843,0.627450980392157,0.172549019607843}
\definecolor{color3}{rgb}{0.83921568627451,0.152941176470588,0.156862745098039}

\begin{axis}[
legend cell align={left},
legend style={fill opacity=0.8, draw opacity=1, at={(0.35,0.35)}, text opacity=1, draw=white!80!black},
ticks=none,
x grid style={white!69.0196078431373!black},
xmin=1000, xmax=1060,
xtick style={color=black},
xticklabels={,,},
xmajorgrids,
y grid style={white!69.0196078431373!black},
ymin=970, ymax=1020,
yticklabels={,,},
ymajorgrids
]
\addplot [semithick, color0, mark=*, mark size=2, mark options={solid}]
table {%
1033.59462199669 995.456910780487
1035.17294284815 996.684314296161
1036.85536226684 997.885206048911
1038.66927263477 999.037583796093
1040.65092605325 1000.12238188012
1042.84331371867 1001.11728911821
1045.33398020516 1002.00971140469
1048.31174459088 1002.80122122814
};
\addlegendentry{Vehicle 1}
\addplot [semithick, color1, mark=*, mark size=2, mark options={solid}]
table {%
1038.8816657769 989.370064057131
1038.00049555909 990.60403070859
1037.14834939289 991.892362384563
1036.30266842998 993.261168748972
1035.4600502272 994.704052288486
1034.61230892451 996.213926500219
1033.7452039576 997.782543581874
1032.83757902958 999.399180413525
};
\addlegendentry{Vehicle 2}
\addplot [semithick, color2, mark=*, mark size=2, mark options={solid}]
table {%
1035.43516955864 982.060890399363
1035.3434875252 983.355105401121
1035.28881649371 984.717432920605
1035.26637817099 986.148066976345
1035.27101098383 987.647229647278
1035.29862015613 989.211431006407
1035.34775124087 990.833622043909
1035.42017349382 992.507385707357
};
\addlegendentry{Vehicle 3}
\addplot [semithick, color3, mark=*, mark size=2, mark options={solid}]
table {%
1025.86573053947 1011.00469164913
1024.20872743956 1011.14579803254
1022.54476172923 1011.19314657044
1020.87992318087 1011.13422900105
1019.22195859504 1010.96109833427
1017.57921471362 1010.67091981858
1015.95951211375 1010.26584684191
1014.3691366837 1009.75237543161
};
\addlegendentry{Vehicle 4}
\end{axis}

\end{tikzpicture}

%% file: figs/VelocityProf.tex
\begin{tikzpicture}

\definecolor{color0}{rgb}{0.12156862745098,0.466666666666667,0.705882352941177}
\definecolor{color1}{rgb}{1,0.498039215686275,0.0549019607843137}
\definecolor{color2}{rgb}{0.172549019607843,0.627450980392157,0.172549019607843}
\definecolor{color3}{rgb}{0.83921568627451,0.152941176470588,0.156862745098039}

\begin{axis}[
legend cell align={left},
legend style={
  fill opacity=0.8,
  draw opacity=1,
  text opacity=1,
  at={(0.02,0.2)},
  anchor=north west,
  draw=white!80!black,
  legend columns=3
},
tick align=inside,
tick pos=left,
xmajorgrids, ymajorgrids,
x grid style={white!69.0196078431373!black},
xlabel={Time (s)},
xmin=-0.1, xmax=1.6,
xtick style={color=black},
y grid style={white!69.0196078431373!black},
ylabel={Velocity (m/s)},
ymin=4.5, ymax=13,
ytick style={color=black}
]

\addplot [ultra thick, black, dashed, mark options={dashed}]
table {%
-.1 11.17
1.6 11.17
};
\addlegendentry{$v_\text{target}$}
\addplot [semithick, color0, mark=*, mark size=2, mark options={solid}]
table {%
0 9.94493352
0.2 10.2449322439756
0.4 10.54493181013
0.6 10.8449318546308
0.8 11.1142860397008
1 11.1604240732392
1.2 11.168382770885
1.4 11.1699236532621
};
\addlegendentry{Vehicle 1}
\addplot [semithick, color1, mark=*, mark size=2, mark options={solid}]
table {%
0 7.74348089
0.2 7.64966974752699
0.4 7.94966845457127
0.6 8.24966659243806
0.8 8.54966494108171
1 8.84966349890434
1.2 9.14966226450472
1.4 9.44966123667958
};
\addlegendentry{Vehicle 2}
\addplot [semithick, color2, mark=*, mark size=2, mark options={solid}]
table {%
0 6.5109571
0.2 6.81095710747544
0.4 7.11095711479875
0.6 7.41095712182094
0.8 7.71095712839918
1 8.01095713439972
1.2 8.31095713970057
1.4 8.61095714419403
};
\addlegendentry{Vehicle 3}
\addplot [semithick, color3, mark=*, mark size=2, mark options={solid}]
table {%
0 8.3942871
0.2 8.39428710721355
0.4 8.39428711428036
0.6 8.39428712105667
0.8 8.39428712740466
1 8.39428713319524
1.2 8.39428713831068
1.4 8.394287142647
};
\addlegendentry{Vehicle 4}
\end{axis}

\end{tikzpicture}

%% file: datatable.tex
\begin{tabular}{@{}r RRRR @{}}
\toprule

\textbf{Model} &  
{\textbf{Collisions}}  & {\mathbf{v_\text{target}-v_\text{avg}}} & {\textbf{\# Players}} & {\textbf{Time (s)}}\\
\midrule
\texttt{IDM}            & 55.80 \scriptstyle\pm 11.53 & 8.68 \scriptstyle\pm 0.13  & 1\phantom{\scriptstyle\pm 0.000 } & 0.005\scriptstyle\pm 0.001 \\
\texttt{C-Nash}         & 4.80 \scriptstyle\pm 1.88 & 3.22 \scriptstyle\pm 0.16& 6.23 \scriptstyle\pm 2.34& 6.43 \scriptstyle\pm 5.74 \\
\texttt{Dec-Nash}           & \mathbf{0.20 \scriptstyle\pm 0.20}& \mathbf{2.79 \scriptstyle\pm 0.12}& \mathbf{1.83 \scriptstyle\pm 1.00} & 2.24 \scriptstyle\pm 1.97\\
\bottomrule
\end{tabular}

%% file: 5-conclusion.tex
\section{Conclusion} \label{sec:conclusion}

Generating safe, human-like behavior in dense traffic is challenging for rule-based planners because of the complex negotiations which drivers undergo to resolve conflicts. Game-theoretic planning shows promise in allowing multiple agents to safely navigate a dense environment because it enables agents to reason about how their plans affect other agents' plans. Though finding a Nash equilibrium in a general-sum Markov game can be quite expensive, recent work has demonstrated significant improvements in optimization efficiency for games with nonlinear dynamics and objectives \citep{fridovich2020efficient, cleac2019algames}. These methods, however, are shown empirically to scale poorly with the number of agents in the game.

In this project, we take a decentralized approach to game-theoretic planning for urban navigation, in which we assume that planning games are only played between vehicles that are mutually interacting (e.g. who are visible to one another within a strongly connected graph component). This assumption significantly reduces the number of agents in each game, and therefore collision constraints, allowing reduced Nash equilibrium optimization time. We demonstrate our approach in roundabouts, where we control agent accelerations assuming paths are known a priori (e.g. from a trajectory planner). We observe the emergence of human-like driving behavior which is safe and efficient when compared to a baseline IDM adaptation, and which is quick when compared to centralized game-theoretic planning.

Our approach has the limitation that it assumes shared knowledge of other agents' objective functions within a strongly connected component, and is therefore not fully decentralized. Future work involves relaxing this assumption. Recent work tries to infer parameters of other players' unknown objective functions~\citep{cleac2020lucidgames}, but this presents challenges in the decentralized case when other players may only stay in a strongly connected component for a short period of time. Future work also involves designing objectives to better imitate human behavior.